\documentclass[9pt]{article}

\usepackage{spconf,amsmath,graphicx}
\usepackage[utf8]{inputenc}
\usepackage[english]{babel}
\usepackage{xcolor}
\usepackage{booktabs}
\usepackage{siunitx}
\usepackage{lineno}
\usepackage{float}
\usepackage{INTERSPEECH2022}

\newcommand{\boldentry}[2]{%
  \multicolumn{1}{S[table-format=#1,
                    mode=text,
                    text-rm=\fontseries{b}\selectfont
                   ]}{#2}}

\title{Transformer-Based Video Front-Ends for Audio-Visual Speech Recognition for Single and Multi-Person Video}

\name{Dmitriy Serdyuk \qquad Otavio Braga \qquad Olivier Siohan}
\address{Google, 111 8th Ave, New York, 10011 USA}
         
\email{\{dserdyuk,obraga,siohan\}@google.com}

\usepackage{amsmath,amsfonts,bm}

\def\eqref#1{equation~\ref{#1}}

\def\1{\bm{1}}

\DeclareMathAlphabet{\mathsfit}{\encodingdefault}{\sfdefault}{m}{sl}
\SetMathAlphabet{\mathsfit}{bold}{\encodingdefault}{\sfdefault}{bx}{n}
\newcommand{\tens}[1]{\bm{\mathsfit{#1}}}
\def\tA{{\tens{A}}}

\def\tF{{\tens{F}}}

\def\tP{{\tens{P}}}

\def\tV{{\tens{V}}}

\newcommand{\R}{\mathbb{R}}

\begin{document}
\topmargin=0mm
\ninept

\maketitle

\begin{abstract}
Audio-visual automatic speech recognition (AV-ASR) extends speech recognition by introducing the video modality as an additional source of information.
In this work, the information contained in the motion of the speaker's mouth is used to augment the audio features.
The video modality is traditionally processed with a 3D convolutional neural network (e.g. 3D version of VGG).
Recently, image transformer networks~\cite{Dosovitskiy2020-nh} demonstrated the ability to extract rich visual features for image classification tasks.
Here, we propose to replace the 3D convolution with a video transformer to extract visual features.
We train our baselines and the proposed model on a large scale corpus of YouTube videos.
The performance of our approach is evaluated on a labeled subset of YouTube videos as well as on the LRS3-TED public corpus.
Our best video-only model obtains 31.4\% WER on YTDEV18 and 17.0\% on LRS3-TED, a 10\% and 15\% relative improvements over our convolutional baseline.
We achieve the state of the art performance of the audio-visual recognition on the LRS3-TED after fine-tuning our model (1.6\% WER).
In addition, in a series of experiments on multi-person AV-ASR, we obtained an average relative reduction of 2\% over our convolutional video frontend.
\end{abstract}

\begin{keywords}
Audio-visual speech recognition, lip reading, video transformer, deep learning.
\end{keywords}

\section{Introduction}
\label{sec:intro}

Many real-world applications of speech recognition operate on a video input (e.g. YouTube videos, webcasts, internet streams, TV broadcasts).
\emph{Audio-visual automatic speech recognition}~(AV-ASR,~\cite{Neti2000-ca,Gupta2017-lz,Makino2019-lm}) adds the video modality to the traditional speech recognition. It has been shown that the video may help recognition, especially in adverse audio conditions~\cite{Makino2019-lm}. The extreme case occurs when the audio is unavailable, a scenario known as~\emph{lip reading}~\cite{Afouras2019-jp}.

A typical end-to-end system for AV-ASR requires a strong visual feature extractor -- the \emph{video front-end}.
This critical component of the AV-ASR system encodes the movements of the speaker's lips movements into the features used downstream for recognition.
Usually, the video front-end is a trainable 3D convolutional network (e.g. a 3D variant of VGG,~\cite{Simonyan2015-tq}).

In order to improve the video front-end, we draw inspiration from the recent works in the area of NLP.
A self attention-based~\cite{Bahdanau2014-vc} \emph{transformer}~\cite{Vaswani2017-di} architecture was proposed for a variety of sequential tasks.
The transformer was instrumental for developing strong NLP~\cite{Devlin2018-gz, Radford2019-wr} and ASR~\cite{Zhang2020-nr} systems.
In the area of computer vision the convolutional networks are the model of choice for image processing.
Recently, it has been shown that a transformer architecture (\emph{vision transformers},~ViT,~\cite{Dosovitskiy2020-nh}) is viable for image classification.
The proposed transformer is able to achieve parity or superior performance to the convolutional networks.
Later, this work was extended to the video classification~\cite{Arnab2021-mq, Bertasius2021-ke}.

Inspired by the success of vision transformers, we propose to use a transformer-based architecture for the video front-end of the AV-ASR system.
We design a video transformer front-end which takes a sliding 3D window of the video.
This window is split into 3D patches of the size 32x32x8 pixels.
Then following~\cite{Dosovitskiy2020-nh, Arnab2021-mq}, we apply an affine transform to the patches followed by a transformer encoder.

This work extends our workshop paper~\cite{Serdyuk2021}.
We add the experiments with a stronger conformer~\cite{Gulati2020-jh} encoder.
We conduct extra experiments on the LRS3-TED with the artificially added noise.

Compared to our previous work in~\cite{Serdyuk2021}, the contributions of this paper are:
\begin{itemize}
    \item We test the feasibility of the transformer-based video front-end for the AV-ASR and lip reading. We design a model that uses an off-the-shelf transformer for the video encoding.
    \item We experimentally evaluate the proposed model. We train on a large scale dataset of YouTube videos.  We experiment with the transformer encoder and the conformer audio-visual encoders. We evaluate on the YTDEV18 and LRS3-TED datasets. The experiments show that the transformer video front-end works at least as good as the convolution. Moreover, we achieve the state of the art performance on the LRS3-TED dataset after fine-tuning our model.
    \item We investigate the proposed model for the multi-person data. Our model outperfroms the convolution baseline.
\end{itemize}

\section{Related Work}
\label{sec:related}

\paragraph*{Audio-Visual Automatic Speech Recognition.}
Audio-visual speech recognition~\cite{Neti2000-ca} made significant
progress thanks to the introduction of the end-to-end approaches~\cite{Assael2016-zg, Chung2016-bd, Makino2019-lm}.
Using deep neural networks and end-to-end training allowed these and other works
to tackle audio-visual speech recognition ``in the wild'',
i.e. unconstrained open-world utterances.

Recently, a remarkable work~\cite{Ma2021-al} achieved the state of the art on the tasks
of audio-visual speech recognition and lip reading by combining CTC~\cite{Graves2006-vf} and seq2seq~\cite{Bahdanau2014-vc} losses with a conformer network~\cite{Gulati2020-jh}.

\vspace{-0.5cm}
\paragraph*{Transformer-based Models for Video.}
Since the attention-based~\cite{Bahdanau2014-vc} transformer architecture was introduced in~\cite{Vaswani2017-di},
it quickly became a model of choice for natural language processing~\cite{Devlin2018-gz, Radford2019-wr}.
Later, it was employed for other sequence modeling tasks, such as speech recognition~\cite{Zhang2020-nr}.
A highly influential paper on \emph{visual transformers}, ViT,~\cite{Dosovitskiy2020-nh}, was the first work demonstrating that
the transformer architecture performs at least as well as convolutions.

Next, the visual transformer was extended from still images to video~\cite{Arnab2021-mq, Sharir2021-mh, Neimark2021-md}, in particular for tasks such as video classification and action recognition.
In contrast to these works, this paper focuses on a sequence-to-sequence task.
While the sequence output provides a signal stronger than classification,
the task is harder due to the fact that the network is required to learn
the alignment.

\section{Model}
\label{sec:model}

\begin{figure}
    \centering
    \includegraphics[width=0.3\textwidth]{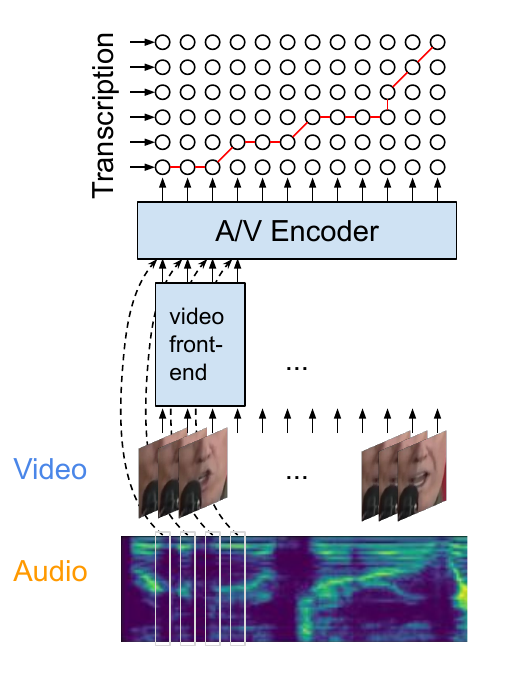}
    \caption{An overview of end-to-end AV-ASR and lip reading models.
             The video is encoded with a video front-end.
             The visual features and the acoustic features are concatenated
             and fed through the AV encoder to be used for the RNN-T loss.}
    \label{fig:avasr}
\end{figure}

In this section we outline the general pipeline for the audio-visual ASR used in our experiments.
Then, we describe the video front-ends, which is a focus of this paper.
We introduce the baseline convolutional front-ends and the proposed transformer-based video front-end.

\subsection{Common A/V ASR Model Architecture}

The common AV-ASR pipeline (Fig.~\ref{fig:avasr}) is shared between all of our experiments.

\vspace{-0.5cm}
\paragraph*{Acoustic Features.} We extract 80 $\log$ Mel filterbank features from the 16kHz input signal with a 25ms wide Hann window with steps of 10ms.
Then, we fold each 3 consecutive features to produce 240-dimensional input which we denote as $\tA \in \R^T \times \R^{D_a}$, where $T$ is the number of time-steps, $D_a = 240$ is the dimensionality of the acoustic features. This corresponds to the acoustic features with the frequency of $\approx33.3$Hz.

\vspace{-0.5cm}
\paragraph*{Visual Features.} The source videos have varying frame rates in the range from 23 to 30 frames per second.
In order to synchronize the features, we re-sample the video frames at the frequency of the audio features (33.3Hz) using the nearest neighbor interpolation.
Then, we crop the videos centering near the mouth region to produce frames of the size $128 \times 128$.
The video is fed then into the video front-end (Section~\ref{ssec:model:video-front-ends}) which yields the video features $\tV \in \R^{T} \times \R^{D_v}$ of dimension $D_v = 512$.

\vspace{-0.5cm}
\paragraph*{Encoder.}
The encoder combines two modalities and embeds them for the use in the decoder.
In all our audio-visual experiments we concatenate the 240-dimensional audio features with the 512-dimensional video features to produce the fused features $\tF = [\tA; \tV] \in \R^T \times \R^{D_a + D_v}$, totalling 752 input features at each time step.
For the video-only (lip reading) we ignore the audio features and decrease the input dimension of the encoder (512 input features), which is $\tF = \tV$.
We use two architectures for the encoder:
\begin{enumerate}
    \item{Transformer:} a 14 layer transformer encoder~\cite{Vaswani2017-di} with 512 hidden dimensions, 8 attention heads, and the relative positional embedding. The self-attention window is limited to $100$ timesteps on left and right.
    \item{Conformer:} a conformer encoder~\cite{Gulati2020-jh} with $17$ layers. The hidden dimension is again 512 and the kernel size of 32.
\end{enumerate}

\vspace{-0.5cm}
\paragraph*{Decoder.} The decoder is a two layer LSTM network, where each layer has 2048 units.
The RNN-T~\cite{Graves2012-ad} loss produces the character level output.

\subsection{Video Front-Ends}
\label{ssec:model:video-front-ends}

The pre-processed video is fed into a video front-end.
In this work we use two types of the video front-ends: the baseline (2+1)D convolutional network and the video transformer network.

\subsubsection{(2+1)D ConvNet Baseline}
\label{ssec:model:baseline}

Our baseline system uses a VGG 3D front-end~\cite{Makino2019-lm} with the following change.
We decompose each filter into the spatial and temporal dimensions.
For example, a $[3, 3, 3]$ filter becomes two filters $[1, 3, 3]$ and $[3, 1, 1]$.
This modification reduces the memory requirements and regularizes the model.
In all our experiments we use a 10 layer convolutional network\footnote{Kernel sizes are: 23, 64, 230, 128, 460, 256, 921, 512, 460, 512.}.
We refer to this baseline as the VGG (2+1)D.

\subsubsection{Video Transformer Front-end}
\label{ssec:model:vit}

\begin{figure*}[h]
    \centering
    \includegraphics[width=\textwidth, trim=0 350 0 100,clip]{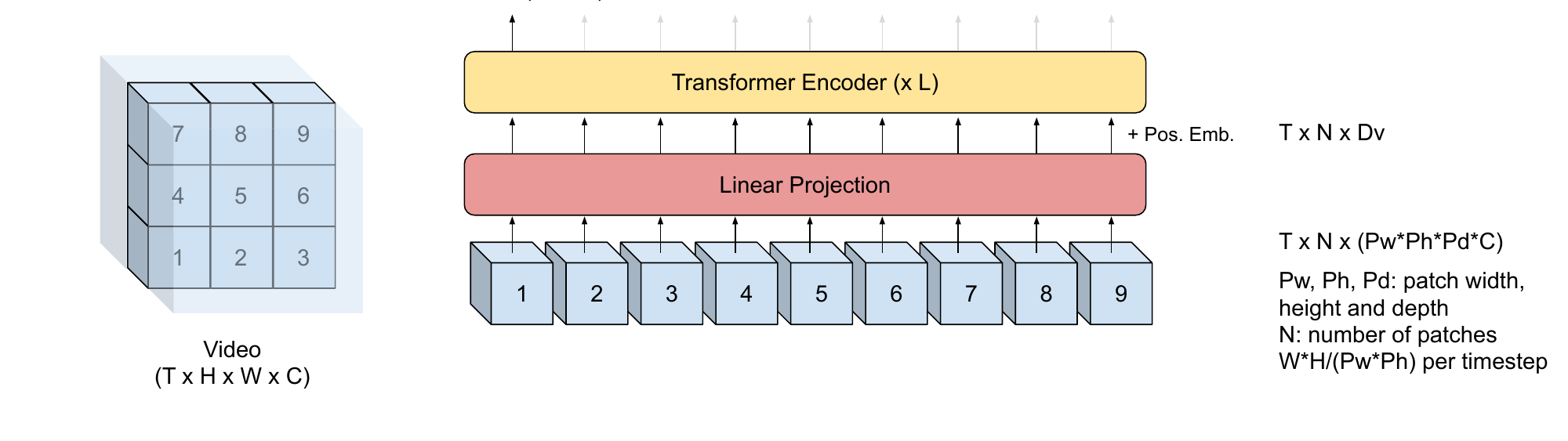}
    \caption{An overview of the proposed architecture for the video-encoding transformer.
             The input video is split into `tubelets'.
             The tubelets are embedded with a linear projection and fed into a transformer.}
    \label{fig:avasr_vit}
\end{figure*}

One of the goals of our design of the video transformer is to reuse the available implementations of the transformer architecture.
We aim to use the minimal number of modifications for the existing models.
This model is visualized in the Fig.~\ref{fig:avasr_vit}.


First, we extract patches of size $P_w \times P_h$ ($= 32 \times 32$) resulting into $N = W / P_w \times H / P_h$ ($= 4 \times 4= 16$) patches at each timestep of the video.
Then, we fold each $P_d$ ($= 8$) consecutive frames which yields a series of \emph{tubelets} $\tP_c \in \R^{P_w} \times \R^{P_h} \times \R^{P_d}$ (3D version of a patch).
After repeating this for each channel and each timestep, we flatten the tubelet tensor $\tP \in \R^{T} \times \R^{N} \times \R^{P_w} \times \R^{P_h} \times \R^{P_d} \times \R^{C} \rightarrow  \tP^{flat} \in \R^{T} \times \R^{N} \times \R^{P_w \times P_h \times P_d \times C}$.
Next, the flattened tensor of patches is fed into an affine transform shared in between patches and the timesteps: $W_{kl} \tP^{flat}_{ijk} + b_k$ which yields a sequence of features $\tP^{feats} \in \R^T \times \R^N \times \R^{D_v}$ ($D_v = 512$).

The transformed tensor of patches is combined with relative positional embeddings (we found that the relative and the absolute positional embeddings have the same performance).
Finally, we run a transformer encoder over the $N$ dimension treating the $T$ as the batch dimension.
We use a 6 layer transformer encoder~\cite{Vaswani2017-di}.
Each layer consists of the self-attention and the feed-forward layer, which are combined with the layer-norm and a residual connection.

Finally, we take the first output of the transformer encoder and discard the rest thus producing a set of the video features $\tV \in \R^{T} \times \R^{D_v}$.

\section{Experiments}
\label{sec:exp}

In this section we describe our experiments and report the results.
We started by training a lip reading system (Section~\ref{ssec:exp:lip_reading}) with a transformer and conformer encoders.
In both cases the proposed video transformer front-end outperforms the convolutional baseline.
Then, we train an audio-visual model using both input modalities.
We find that the video transformer matches the baseline or performs slightly better than the baseline.

\vspace{-0.5cm}
\paragraph*{Datasets.} We train on a dataset mined from public YouTube videos.
We use a semi-supervised procedure proposed in~\cite{Liao2013-em} and adapted to include videos in~\cite{Makino2019-lm, Shillingford2019-sc}.
This procedure extracts short segments of the video where the force-aligned user uploaded transcript matches the production quality ASR system with high confidence.
Then, only the segments are kept where the video track matches the audio with high confidence.
The resulting dataset contains about 90k hours of transcribed video segments limited to 512 frames (15 seconds).

A separate set of YouTube videos is used for the development and eval sets.
These videos were transcribed by professionals -- the YTDEV18 set~\cite{Makino2019-lm}.
In order to compare to prior publications, we use the LRS3-TED~\cite{Afouras2018-pq} eval set.

\vspace{-0.5cm}
\paragraph*{Training.} In all our experiments we use the batch size 8 and the Adam~\cite{Kingma2014-dh} optimizer ran synchronously on 128 accelerators (totalling the batch size of 1024).
We use the multi-style training (MTR,~\cite{Cui2015-ox}), which increases robustness to the noise.

The transformer models were trained with the following learning rate schedule.
First, the learning rate linearly warmed up to $1e^{-4}$ for the first $30,000$ iterations.
Then, it is constant until iteration $200,000$.
Finally, it is annealed exponentially down to $1e^{-6}$ until iteration $300,000$.

The conformer models use the learning rate schedule which warms up linearly and then anneals exponentially.
The maximum learning rate is $1.7e^{-2}$ and the number of warm up iterations is $15,000$.

\subsection{Lip Reading}
\label{ssec:exp:lip_reading}

We summarize our findings for the lip reading models in Table~\ref{tab:lip_reading}.
The proposed ViT 3D model model outperforms the VGG baseline when using the transformer encoder (4\% relative improvement on YTDEV18 and 8\% on LRS3-TED) and the conformer encoder (10\% improvement on YTDEV18 and 9\% on LRS3-TED).
Furthermore, our models outperform the previous publications~\cite{Makino2019-lm, Ma2021-al, Afouras2018-gl} with a caveat that we use a different training set.

From these experiments we conclude that the ViT is able to provide strong visual features for the lip reading task.

\subsection{Audio-Visual Automatic Speech Recognition}
\label{ssec:exp:av_asr}

The experiments on the combined audio and video follow the same protocol as the lip reading with an exception that the encoder concatenates the audio features to the extracted video features.
Then we artificially add the audio noise to the YTDEV18 (following~\cite{Makino2019-lm}).
We add the babble noise of signal-to-noise ratios 20dB, 10db, and 0dB.
The noise was drawn from the NoiseX database~\cite{varga1993assessment}.
Then, we overlap a fixed random utterance from the test set (denoted ``Overlap''). 
The results are summarized in Table~\ref{tab:av_asr}.

We find that our ViT front-end matches the performance of the VGG baseline.
When using the transformer encoder we observe a slight increase in the performance for 0dB noise condition.

The lower part of the table refers to a stronger conformer encoder.
The ViT front-end is able to match the convolutional front-end.

For both the transformer and conformer encoders we see a small dip in the performance for the overlap noise.
One of possible reasons for this is that the model is trained with MTR which mitigates the babble noise but not the overlap noise.
Notice that the system is still able to improve upon the video-only setup in Table~\ref{tab:lip_reading} (29.9\% vs 31.4\%).

\subsection{Multi-Person Audio-Visual Recognition}

In this section, we briefly summarize our model applied to multi-person A/V ASR. 
We closely follow the setup in~\cite{Braga2020-yw,Braga2021-lj}, where multiple videos are encoded with the shared video front-end followed by the attention to choose the active speaker.
The evaluation sets were artificially constructed by mixing the existing sets.
The audio and video is taken from one utterance, then several (2, 4, or 8) video distractors are added.

\begin{figure}
    \centering
    \includegraphics[scale=0.25]{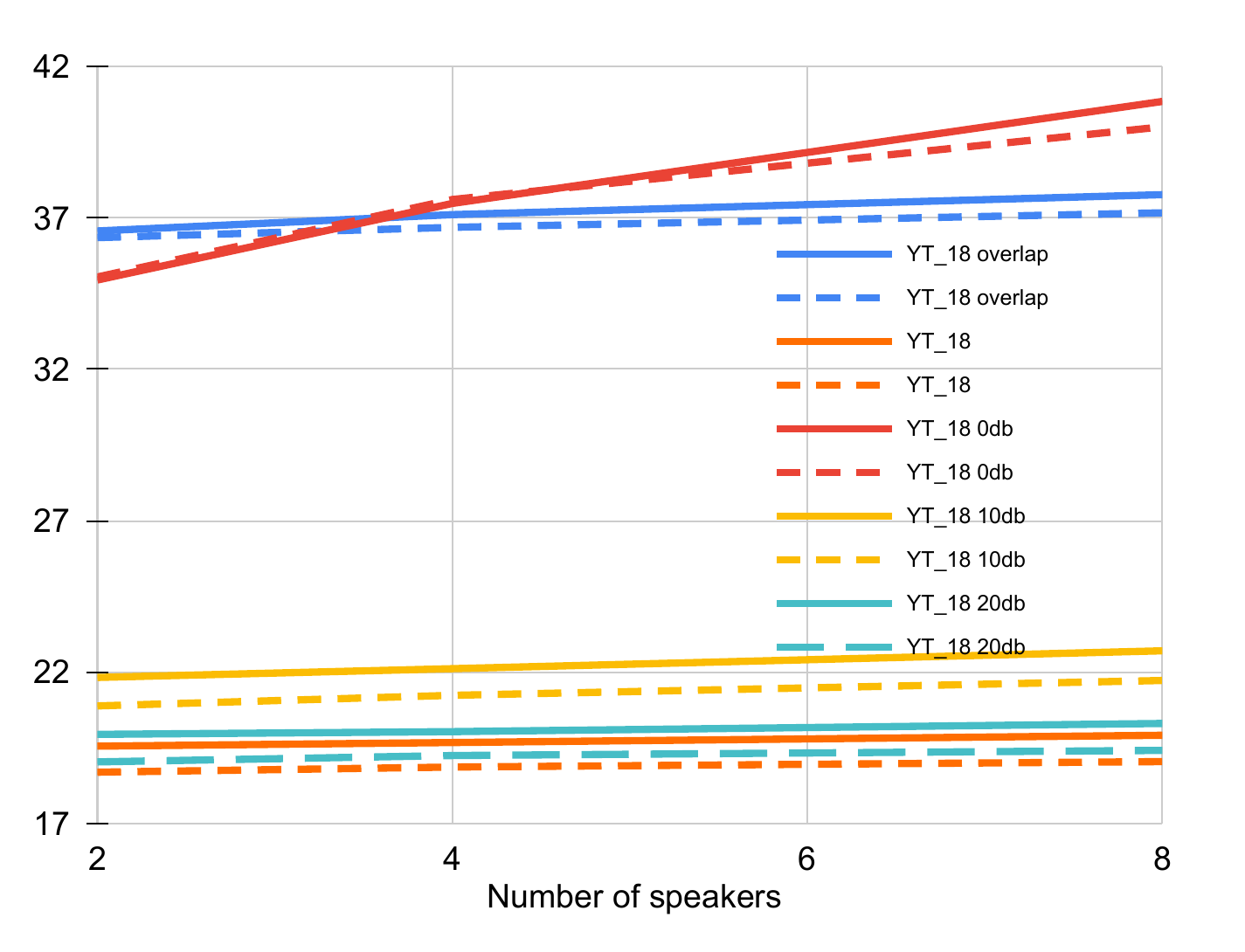}
    \caption{WER for multi-person recognition on YT\_18 data.}
    \label{fig:wer_multiperson}
    \vspace{-0.7cm}
\end{figure}

We report the WER results with a varying number of video distractor speakers in the Figure~\ref{fig:wer_multiperson}.
The solid lines are the baseline VGG video front-end, and the dash lines stand for the proposed ViT front-end.
Our model outperforms the baseline in the majority of conditions.

\subsection{Audio-Visual Recognition on LRS3-TED and Fine-Tuning}

The LRS3-TED tends to have higher audio quality compared to the majority of the YouTube videos we use for training. 
In order to close this gap, we fine-tune our models on the LRS3-TED training set.
More specifically, we train our models for $10,000$ steps on a 50-50 mix of the YouTube and the LRS3-TED training data.
We use the maximum learning rate of $1e^{-5}$ which was warmed up linearly from 0 across the first 200 steps and then held constant.

\begin{table}[t]
    \centering
    \caption{Lip-reading performance, \%WER. The proposed model (ViT 3D) outperforms the convolutional baseline (VGG) for both the transformer and conformer encoders.}
    \begin{tabular}{lS[table-format=2.2]S[table-format=2.2]}
        \toprule
        \bfseries Model      & {\bfseries YTDEV18} & {\bfseries LRS3-TED} \\ 
        \midrule
        TM-seq2seq~\cite{Afouras2018-gl} & {--} & 58.9 \\
        ResNet+Conf~\cite{Ma2021-al}     & {--} & 43.3 \\
        RNN-T~\cite{Makino2019-lm}       & 48.5 & 33.6 \\
        \midrule
        Transformer encoder: \\
        \hspace{0.2cm} VGG (2+1)D & 40.5 & 28.2 \\ 
        \hspace{0.2cm} ViT 3D     & 38.8 & 25.9  \\ 
        \midrule
        Conformer encoder: \\
        \hspace{0.2cm} VGG (2+1)D & 34.9 & 20.0 \\
        \hspace{0.2cm} ViT 3D & \boldentry{2.2}{31.4} & \boldentry{2.2}{17.0} \\
        \bottomrule
    \end{tabular}
    \label{tab:lip_reading}
\end{table}

\begin{table}[t]
    \centering
    \caption{Audio-visual ASR performance, \%WER. ($\infty$dB) is the clean subset;
             20db, 10dB, 0dB -- data with artificial noise added;
             ``Overlap'' -- contains overlapped utterances.
             The proposed ViT model matches the VGG baseline.
             }
    \begin{tabular}{lccccc}
        \toprule
        \bfseries Model      & $\infty$dB & 20dB & 10dB & 0dB & Overlap \\
        \midrule
        Audio-only & 16.5 & 17.0 & 19.8 & 42.9 & 35.0 \\
        \midrule
        Transformer: \\
        \hspace{0.2cm} VGG (2+1)D  & 14.4 & 14.5 & 15.6 & 23.4 & 31.2 \\
        \hspace{0.2cm} ViT 3D      & 14.4 & 14.6 & 15.6 &  23.1 & 31.9 \\
        \midrule
        Conformer: \\
        \hspace{0.2cm} VGG (2+1)D  & 13.6 & 13.7 & 14.5 & 19.3 & 29.3\\
        \hspace{0.2cm} ViT 3D      & 13.4 & 13.5 & 14.3 & 19.3 & 29.9 \\
        \bottomrule
    \end{tabular}
    \label{tab:av_asr}
    \vspace{-0.7cm}
\end{table}

The results for A/V are reported in Table~\ref{tab:ted_results}.
The fine-tuned transformer model matches the previous state of the art~\cite{Ma2021-al} for supervised models.
The conformer-based models are reported in the lower section of Table~\ref{tab:ted_results}.
We found that the audio signal is strong enough to achieve the WER of 1.6\%.
Both the baseline AGG AV-ASR and the proposed AV ViT models match this result which demonstrates that the performance on TED is nearly saturated.
Therefore, we corrupt the LRS-TED test set by adding the babble noise of 20dB, 10dB, and 0dB.
The performance of the audio only model rapidly drops down to 6.1\% for 0dB noise.
In comparison, the AV models score 3.1\% and 2.9\% for the VGG and ViT front-ends.

As a side note, we were surprised that all the tested models performed so well in the presence of the 0dB babble noise (compare this to the performance drop in Table~\ref{tab:av_asr}).
The main reason for this is the very high quality of the audio.

Finally, we did not observe any benefit in fine-tuning our lip reading models on LRS3-TED.
We hypothesise that the domain shift between our train data and the LRS3-TED test set is the greatest for the audio modality (clean, professionally recorded audio).



\begin{table}[t]
    \centering
    \caption{AV-ASR performance on the LRS3-TED dataset.
             Models denoted with $^*$ are trained on a large dataset of YouTube videos.
             $^\dagger$ denotes self-supervised pre-training.
             All conformer models are fine-tuned for the TED data.}
    \label{tab:ted_results}
    \begin{tabular}{lS[table-format=2.1]S[table-format=2.1]S[table-format=2.1]S[table-format=2.1]}
         \toprule
         \bfseries Model & {WER, \%} & {20dB} & {10dB} & {0dB}  \\
         \midrule
         TM-CTC~\cite{Afouras2018-gl}   & 27.7 & {--} & {--} & {--} \\
         EG-s2s~\cite{Xu2020-sf}        & 6.8 & {--} & {--} & {--} \\
         RNN-T~\cite{Makino2019-lm}$^*$ & 4.5 & {--} & {--} & {--} \\
         ResNet+Conf~\cite{Ma2021-al}   & 2.3 & {--} & {--} & {--} \\
         AV-Hubert~\cite{Shi2022-us}$^\dagger$    & \boldentry{2.1}{1.3} & {--} & {--} & {--} \\
         \midrule
         Transformer encoder: \\
         \hspace{0.2cm} VGG (2+1)D$^*$  & 3.3 & 3.3 & 3.4 & 6.4 \\
         \hspace{0.2cm} ViT 3D$^*$      & 3.3 & 3.3 & 3.3 & 6.2 \\
         \hspace{0.4cm} + fine-tune     & 2.3 & 2.4 & 2.6 & 5.1 \\
         \midrule
         Conformer encoder: \\
         \hspace{0.2cm} audio-only$^*$  & \boldentry{2.1}{1.6} & 1.5 & 1.8 & 6.1 \\
         \hspace{0.2cm} VGG (2+1)D$^*$  & \boldentry{2.1}{1.6} & 1.7 & 1.8 & 3.1 \\
         \hspace{0.2cm} ViT 3D$^*$      & \boldentry{2.1}{1.6} & 1.6 & 1.7 & \boldentry{2.1}{2.9} \\
         \bottomrule
    \end{tabular}
\end{table}


\vspace{-0.5cm}
\section{Conclusions}
\label{sec:conclusions}

We compared a transformer-based front-end for video encoding to the convolutional front-end.
We conclude that the transformer is a promising new architecture that is at least as good as the convolution.
Furthermore, the ViT outperforms the convolutional baseline in certain settings, such as lip reading and the noisy LRS3-TED.

We fine-tuned our models on the public LRS3-TED dataset.
This allowed the state of the art results on this set.
We observed that the proposed ViT model outperforms the convolutional baseline and the audio-only recognition.

We are unable to use the LRS2-BBC dataset due to licence restrictions, which prohibits the dataset use by the private and industry researchers.
Therefore, we cannot directly compare to some of the previously reported results.


\section{Safe AI Principles}
\label{sec:safe_ai}

We are aware of the sensitive nature of the AV-ASR research
and other AI technologies used in this work.
Therefore, we ensure that this work abides by the Google AI Principles~\cite{noauthor_undated-lg}.

\section{Acknowledgements}

We would like to acknowledge support and advice of all our team mates, especially
Hank Liao, Oscar Chang, Basi Garcia, and Kishan Sachdeva.

\bibliographystyle{IEEEbib}
\bibliography{lit}

\end{document}